# On Generalized Rectangular Fuzzy Model for Assessment

Igor Ya. Subbotin

Department of Mathematics and Natural Sciences,
College of Letters and Sciences, National University,
Los Angeles,California, USA
E-mail: isubboti@nu.edu

## Abstract

The author constructs a new Generalized Rectangular Model (GRM) for assessment and discusses its place among other previously developed models (the Rectangular Model, Triangular Model, and Trapezoidal Model). More importantly, a generalized approach unifying all these models and criteria for their applications to assessment were also developed. This generalizes and significantly simplifies the process of the listed models' implementation. A concrete example and supporting analysis based on the application of the fuzzy models to learning assessment are given.

## Academic Discipline and Sub-Disciplines

Mathematics, Fuzzy Sets, Mathematics Education

## SUBJECT CLASSIFICATION

## TYPE (METHOD/APPROACH)

Research article

## INTRODUCTION

The majority of commonly used in practice assessment methods are traditionally based on the principles of the classical, bivalent logic. However, due to the human nature, there are many cases where traditional approach is not the completely suitable. For example, a teacher is frequently not sure about a particular numerical grade he should assign. Fuzzy logic, due to its nature of characterizing a case with multiple values, offers wider and richer resources covering such kind of cases. For general facts on fuzzy sets we refer to the book [1].

Within the last decade, some useful applications based on the fuzzy centroid method (see, for example [2]), have been developed by the author and his collaborators in [3-11] and then implemented in some recent researches (see, for example [12-14]). In the current article we construct a new Generalized Rectangular Model (GRM) for assessment and discuss its place among other previously developed models (the Rectangular Model, Triangular Model, and Trapezoidal Model). More importantly, a generalized approach unifying all these models and criteria for their applications to assessment were also developed. This generalizes and significantly simplifies the process of the listed models' implementation. A concrete example and supporting analysis based on the application of the fuzzy models to learning assessment are given.

## 1. THE RECTANGULAR MODEL

The main idea of the Rectangular Model was developed in [3, 4].

Given a fuzzy subset $A = \{(x, m(x)): x \in U\}$ of the universal set U of the discourse with membership function $m: U$ to $[0, 1]$, we correspond to each x U an interval of values from a prefixed numerical distribution, which actually means that we replace U with a set of real intervals. Then, we construct the graph F of the membership function $y = m(x)$. There is a commonly used in Fuzzy Logic approach to measure performance with the pair of numbers $(x_c, y_c)$ as the coordinates of the centre of gravity (centoid), say Fc, of the graph F, which we can calculate using the following well-known formulas:

$$x_c = \frac{\iint\limits_{F} x\,dx\,dy}{\iint\limits_{F} dx\,dy}, \; y_c = \frac{\iint\limits_{F} y\,dx\,dy}{\iint\limits_{F} dx\,dy}. \qquad (1)$$

We can characterize a performance as very low (F) if x∈[0, 1), as low (D) if x∈ [1, 2), as intermediate (C) if x∈[2, 3), as high (B) if x∈[3, 4), and as very high (A) if x ∈[4, 5] respectively.

Consider a concrete example concerning two classes' performances comparing. Denote by $C_1$ the first class of students and by $C_2$ the second class respectively, and the set  U={A, B, C, D, F} is the set of usual labels-grades of performance ordered in the way of the preferences A > B > C > D > F. We are going to represent the $C_i$, i=1, 2, as fuzzy subsets of U. For this, if $n_iF$, $n_iD$, $n_iC$, niB and $n_iA$ denote the number of students of class $C_i$ who achieved very low, low, intermediate, high, and very high success respectively, we define the membership function $mC_i$ in terms of the frequencies, i.e. by

$mC_i(x)=\dfrac{n_{ix}}{n}$ for each x in U. Thus we can write $C_i$ = {(x, $\dfrac{n_{ix}}{n}$) : x ∈ $U$}, i=1,2.

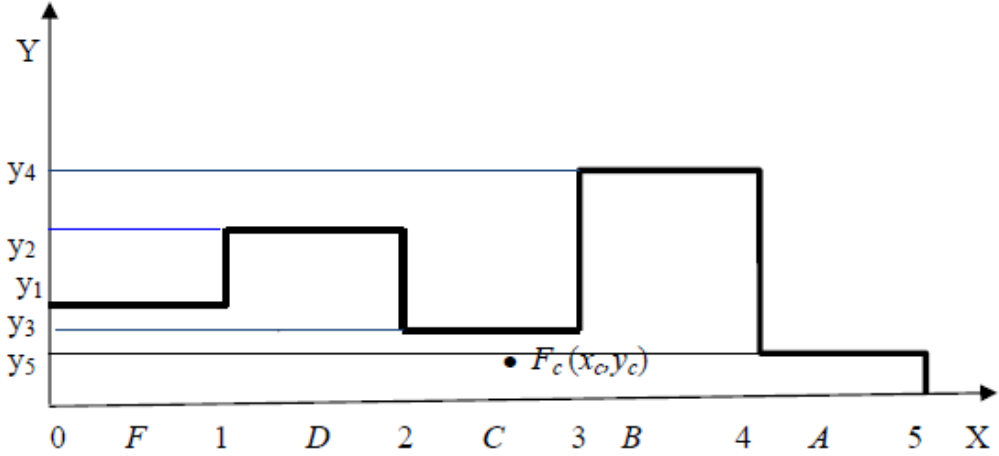

**Figure 1.** Bar graphical data representation for the Rectangular Model

Therefore in this case, the graph F of the corresponding fuzzy subset of U is the bar graph consisting of five rectangles, say Fi, i=1,2,3, 4 , 5 , whose bases lying on the x axis have length 1 (see the diagram 1). In the general case, $\sum_{i=1}^{n} y_i = \iint_F dxdy = 1$ is the area of F, and

$$\iint_F xdxdy = \frac{1}{2}\sum_{i=1}^{5}(2i-1)y_i \text{ and } \iint_F ydxdy = \sum_{i=1}^{n}\int_0^{y_i} ydy = \frac{1}{2}\sum_{i=1}^{n} y_i^2.$$

Therefore formulas (1) can be transformed into the following form:

$$x_c = \frac{1}{2}\left(\frac{y_1 + 3y_2 + 5y_3 + 7y_4 + 9y_5}{y_1 + y_2 + y_3 + y_4 + y_5}\right), \tag{2}$$

$$y_c = \frac{1}{2}\left(\frac{y_1^2 + y_2^2 + y_3^2 + y_4^2 + y_5^2}{y_1 + y_2 + y_3 + y_4 + y_5}\right).$$

Without loss of the generality, we can assume that $y_1+y_2+y_3+y_4+y_5$ = 1. Therefore we can write

$$x_c = \frac{1}{2}(y_1 + 3y_2 + 5y_3 + 7y_4 + 9y_5), y_c = \frac{1}{2}(y_1^2 + y_2^2 + y_3^2 + y_4^2 + y_5^2). \tag{3}$$

with $y_i = \dfrac{m(x_i)}{\sum_{x \in U} m(x)}$ . But $0 \le (y_1 - y_2)^2 = y_1^2 + y_2^2 - 2y_1 y_2$, and therefore $y_i^2 + y_j^2 \ge 2y_i y_j$ with the equality holding if and only if $y_i = yj$.

Hence it is easy to check that $(y_1 + y_2 + y_3 + y_4 + y_5)^2 \le 5(y_1^2 + y_2^2 + y_3^2 + y_4^2 + y_5^2)$, with the equality holding if and only if $y_1 = y_2 = y_3 = y_4 = y_5$. But $y_1 + y_2 + y_3 + y_4 + y_5 = 1$, and therefore

$$1 \leq 5(y_1^2 + y_2^2 + y_3^2 + y_4^2 + y_5^2) \qquad (4)$$

with the equality holding if and only if $y_1 = y_2 = y_3 = y_4 = y_5 = \dfrac{1}{5}$. Then the first formula of (3) gives that $x_c = \dfrac{5}{2}$. $1 \leq 10y_c$, or $y_c \geq$ $\dfrac{1}{10}$. Therefore the unique minimum for $y_c$ corresponds to the centre of gravity $F_m \left( \dfrac{5}{2}, \dfrac{1}{10} \right)$. The ideal case is when $y_1 = y_2 = y_3 = y_4 = 0$ and $y_5 = 1$. Then from formulas (3) we obtain that $x_c = \dfrac{9}{2}$ and $y_c = \dfrac{1}{2}$. Therefore the centre of gravity in this case is the point $F_i \left( \dfrac{9}{2}, \dfrac{1}{2} \right)$. On the other hand, in the worst scenario, $y_1 = 1$ and $y_2 = y_3 = y_4 = y_5 = 0$. Then by the formulas (3), we find that the centre of gravity is the point $F_x \left( \dfrac{1}{2}, \dfrac{1}{2} \right)$.

Based on the above considerations it is logical to formulate our criterion for comparing the groups' performances in the following form:

*Among two or more groups, the group with the largest $x_c$ performs better. If two or more groups have the same $x_c \geq 2.5$, then the group with the higher $y_c$ performs better. If two or more groups have the same $x_c < 2.5$, then the group with the lower $y_c$ performs better.*

.

## 2. THE GENERALIZED RECTANGULAR MODEL

However, the consideration above does not reflect a very frequent situation when the assessor is not sure about the assessing of the marginal performances closed to two adjacent levels. In this situation, the triangular and trapezoidal models for assessment have been developed in [7-9]. The proposed below new modification, which we will call the generalized rectangular model (GRM), also treats this situation. Here we use the rectangular bar graph in which we allow the rectangles have some overlapping intersection. Namely, we allow to any two adjacent rectangles have 30% of their bases belongs to both of them. This way, we cover the situation of uncertainty in assessment of marginal grades described above. In this case, since the marginal individual scores are considered as common parts for the pairs of the adjacent rectangles, it is logical to count these parts twice by placing the ambiguous scores in both adjacent regions. In this case, we represent each one of the five rectangles by its centers of gravity mi, i=1, 2, 3, 4, 5 and we consider the entire area, i.e. the sum of the areas of the five rectangles, as the system of these points-centers. We denote by $y_i$, i=1,...,5 be the percentages of the individuals' whose performance was characterized by F, D, C, B, and A respectively.

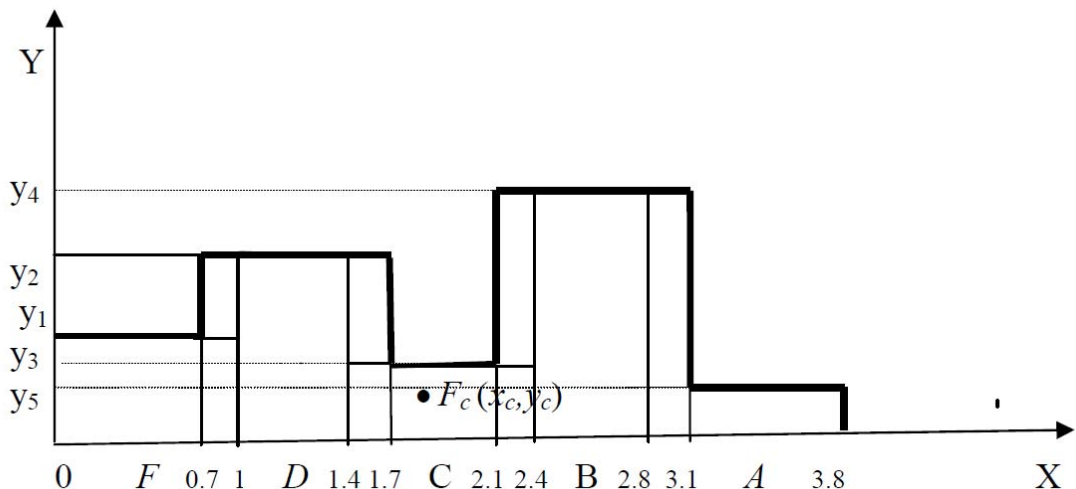

**Figure 2.** Bar graphical data representation for the Generalized Rectangular Model

Since we included the boundary cases to both adjacent rectangles and count their areas twice, we cannot apply here the formulas (1). That is why we use some different approach which gives us a generalization of formulas (2). Note that in the case of the Rectangular Model it will bring us to the same formulas (3). Now, we consider the set of the $m_i$ centres of gravity of rectangles. Since the centre of gravity of a rectangle lies at the intersection of its diagonals, it is easy to see that $m_i$ $(0.7i-0.2, 0.5y_i)$.

We can consider the system of the centres of gravity $m_i$, i=1, 2, 3, 4, 5 and we calculate the coordinates $(x_c, y_c)$ of the centre of gravity $F_c$ of this system by the following general formulas, derived from the commonly used definition (e.g. see [14]):

$$x_c = \frac{1}{S}\sum_{i=1}^{n} S_i x_{c_i}, \; y_c = \frac{1}{S}\sum_{i=1}^{n} S_i y_{c_i}. \tag{5}$$

In formulas (5) $S_i$, i= 1, 2, 3, 4, 5 denote the areas of the corresponding rectangles. Since we decided to count the common parts twice, and since $S_i = y_i$, we obtain $S = \sum_{i=1}^{n} S_i = 1$. Therefore, from formulas (5) we obtain

$$x_c = \sum_{i=1}^{n} y_i(0.7i - 0.2) = 0.7\sum_{i=1}^{n} iy_i - 0.2, \; y_c = \sum_{i=1}^{n} y_i 0.5 y_i = 0.5\sum_{i=1}^{n} y_i^2 \tag{6}.$$

In our particular case n=5, we obtain

$$x_c = 0.7(y_1 + 2y_2 + 3y_3 + 4y_4 + 5y_5), \; y_c = 0.5(y_1^2 + y_2^2 + y_3^2 + y_4^2 + y_5^2) \tag{7}$$

Now we consider the extreme cases. For i, j=1, 2, 3, 4, 5, we have $y_i^2 + y_j^2 \geq 2y_iy_j$, with the equality holding if, and only if, $y_i=y_j$. Therefore

$$1 = (\sum_{i=1}^{5} y_i)^2 = \sum_{i=1}^{5} y_i^2 + 2\sum_{\substack{i,j=1,\\i \neq j}}^{5} y_i y_j \leq \sum_{i=1}^{5} y_i^2 + 2\sum_{\substack{i,j=1,\\i \neq j}}^{5} (y_i^2 + y_j^2) = 5\sum_{i=1}^{5} y_i^2 \; \text{or} \; \sum_{i=1}^{5} y_i^2 \geq \frac{1}{5}, \tag{8}$$

with the equality holding if and only if $y_1 = y_2 = y_3 = y_4 = y_5 = \frac{1}{5}$. In the case of equality, the first of formulas (7) gives that $x_c$ = $0.7(\frac{1}{5} + \frac{2}{5} + \frac{3}{5} + \frac{4}{5} + \frac{5}{5}) - 0.2$ = 2.1-0.2=1.9. Further, combining the inequality (8) with the second of formulas (7) one finds that $y_c \geq 0.1$. Therefore the unique minimum for $y_c$ corresponds to the $F_m(1.9,0.1)$. The ideal case is when $y_1 = y_2 = y_3 = y_4 = 0$ and $y_5 = 1$. Then from formulas (7) we get that $x_c = 3.3$ and $y_c = 0.5$. Therefore the centre of gravity in this case is the point $F_i(3.3, 0.5)$. On the other hand, the worst case is when $y_1 = 1$ and $y_2 = y_3 = y_4 = y_5 = 0$. Then from formulas (7), we find that the centre of gravity is the point $F_w(0.5, 0.5)$.

Based on the above considerations it is logical to formulate our criterion for comparing in GRM the two groups' performance in the following form:

*Between two groups the group with the greater value of $x_c$ demonstrates a better performance. If two groups have the same $x_c \geq 1.9$, then the group with the greater value of $y_c$ demonstrates a better performance. If two groups have the same $x_c < 1.9$, then the group with the smaller value of $y_c$ demonstrates a better performance.*

## 3. IMPORTANT GENERALIZATIONS

Consider all general formulas for all three models:

Generalized Rectangular Model (GRM) : $x_c = 0.7\sum_{i=1}^{n} iy_i - 0.2$, $y_c = 0.5\sum_{i=1}^{n} y_i^2$ ;

Triangular Model (TM): $x_c = 0.7\sum_{i=1}^{n} iy_i - 0.2$, $y_c = 0.2\sum_{i=1}^{n} y_i^2$ [7];

Trapezoidal Model (TrM): $x_c = 0.7\sum_{i=1}^{n} iy_i - 0.2$, $y_c = \frac{3}{7}\sum_{i=1}^{n} y_i^2$ [ 8].

Observe that in all these formulas we are dealing with the same key expressions $\sum_{i=1}^{n} iy_i$ for $x_c$ and $\sum_{i=1}^{n} y_i^2$ for $y_c$. In general, for all three models we deal with the following formulas $x_c = \alpha\sum_{i=1}^{n} iy_i - \beta$, $y_c = \gamma\sum_{i=1}^{n} y_i^2$ where $\alpha$, $\beta$, and $\gamma$ are some coefficients depending on the model shape of the graph we choose. It is not difficult to prove that if we consider the areas of intersection of the bases of the adjacent figures not 30% as in the models above, but any percentage less than 50%, the mentioned key expressions will be hold. So for the comparing purposes it is enough to establish some common criteria based on those expressions. It is easy to formulate it in the following way.

Consider our graph on the interval [0, m] at the x-axes. Then:

*Between two groups the group with the greater value of $x_c$ ($\sum_{i=1}^{n} i y_i$) demonstrates a better performance. If two groups have the same $x_c \geq 0.5m$, then the group with the greater value of $y_c$ ($\sum_{i=1}^{n} y_i^2$) demonstrates a better performance. If two groups have the same $x_c < 0.5m$, then the group with the smaller value of $y_c$ ($\sum_{i=1}^{n} y_i^2$) demonstrates a better performance.*

If the join area between two adjacent figures is *k%*, then $m = n - \dfrac{k}{100}$ *(n-1)*.

In particular in the case when $n = 5$ and $m = 3.8$, we come to the criterion formulated in the previous paragraphs.

How to choose the shape of the model areas? If the data close to the means, the best choice is the Triangular model. If the data distributed uniformly, the choice of rectangles is preferable. The trapezoid shape is the most useful and combined both previous cases as its partial cases (when the upper base of a trapezoid is 0 we come to the Triangular Model, or if it equal to the lower base we come to GRM correspondingly). In any case, the remark above on the key expressions $\sum_{i=1}^{n} i y_i$ for $x_c$ and $\sum_{i=1}^{n} y_i^2$ for $y_c$ shows that if we use the same model the shapes forms are insignificant. But we can use the models not only for comparing in the frame of the same model, but for different purposes.

And the last, but not the least important remark. In the USA system of assessment of learning performance there is a commonly used weighted average is called GPA (Grade Point Average [16]). The class performance here we count using the following formula GPA= $0 y_1 + 1 y_2 + 2 y_3 + 3 y_4 + 4 y_5$ where $y_i$ for i=1, 2, 3, 4, 5 is the percent of students received the grades F, D, C, B, and A correspondingly. Consider our key form in the case of n=5. We have

$$y_1 + 2 y_2 + 3 y_3 + 4 y_4 + 5 y_5 = (y_1 + y_2 + y_3 + y_4 + y_5) + 1 y_2 + 2 y_3 + 3 y_4 + 4 y_5 = 1 + GPA.$$

It shows that comparing two classes *with different GPA*, in the case of n=5, our criteria for all models will bring the same result as comparing GPA's. However, in the case when GPAs coincides, the traditional approach will not lead us to logically based preferences. In this situation, since of its concrete logical nature our criterion becomes useful.

Consider the following example.

**Table 1.** Number of class students reached the following stage of knowledge acquisition

| Grades | Class I | Class II |
|---|---|---|
| F | 0 | 0 |
| D | 0 | 0 |
| C | 10 | 0 |
| B | 0 | 20 |
| A | 50 | 40 |

For both classes the GPA is about 3.7. "The quality of knowledge", i.e. the ratio of the students received B or better to the amount of all students, for the second class is higher than for the first one. The standard deviation for the second class is definitely smaller. So from the common point of view and from the statistical point of view the situation in the second class is better. However, some instructors could prefer the situation in the first class, since there are much more "perfect" students in this class. Everything is determined by the set of goals preference. We choose the GRM model for the analyzing this example (7). Here

$x_{c1}=x_{c2}= 0.7( y_1 + 2 y_2 + 3 y_3 + 4 y_4 + 5 y_5 ) = 0.7(GPA+1)=0.7(4.7)=3.29>1.9.$

Consider $y_c=0.5( y_1^2 + y_2^2 + y_3^2 + y_4^2 + y_5^2 )$. For the first class, $y_{c1} =2600$, which is larger than $y_{c2}=2000$. As we can see and $y_{c1} > y_{c2}$. So, based on our criteria, the first class performs better than the second.